\documentclass[letterpaper, 10 pt, conference]{ieeeconf}  % Comment this line out if you need a4paper

\IEEEoverridecommandlockouts                              %for thanks command
\usepackage{verbatim}
\usepackage{graphics} % for pdf, bitmapped graphics files
\usepackage{amsmath,amssymb,stmaryrd,mathtools, amsfonts}
\usepackage{multirow}% http://ctan.org/pkg/multirow
\usepackage{hhline}
\usepackage{algorithm}
\usepackage[noend]{algpseudocode}
\PassOptionsToPackage{hyphens}{url}\usepackage{hyperref} % break url at hyphens
\usepackage{dsfont}
\usepackage{algorithmicx}
\usepackage{multirow}
\usepackage[utf8]{inputenc}
\usepackage{import}
\usepackage{graphicx, graphics}
\usepackage{subcaption}
\usepackage{array}
\usepackage{color,xcolor}
\usepackage{soul}
\usepackage{booktabs}
\let\labelindent\relax
\usepackage{enumitem}
\definecolor{v0}{rgb}{1, 0.5, 0} % v0 in orange
\definecolor{v1}{rgb}{0, 1, 0.83} % v1 in cyan
\definecolor{v2}{rgb}{0.25, 0.25, 0.25} % v2 in grey
\definecolor{v3}{rgb}{1, 0 ,0.58 } % v3 in magenta

\usepackage{siunitx} %https://tex.stackexchange.com/questions/269841/1-5e-10-style-scientific-notation-looks-ugly-in-latex-math-mode-how-can-i-forma

\usepackage{svg}

\title{\LARGE \bf
Reinforcement Learning and Distributed Model Predictive Control for Conflict Resolution in Highly Constrained Spaces
}

\author{Xu Shen and Francesco Borrelli% 
\thanks{University of California, Berkeley, CA, USA (\{xu\_shen, fborrelli\}@berkeley.edu).}%
}

\begin{document}

\maketitle
\thispagestyle{empty}
\pagestyle{empty}

%%%%%%%%%%%%%%%%%%%%%%%%%%%%%%%%%%%%%%%%%%%%%%%%%%%%%%%%%%%%%%%%%%%%%%%%%%%%%%%%
% \todo{
% Stuff to check in edits
% \begin{itemize}
%     \item ensure consistent wording: ``intent'' vs ``goal''
%     \item consistency of claims in intro and results shown
%     \item clear results/methods - not confusing
%     \item solid motivation and enough lit. review (not too thorough for this paper)
%     \item tense consistency in methods
% \end{itemize}
% }

\begin{abstract}
This work presents a distributed algorithm for resolving cooperative multi-vehicle conflicts in highly constrained spaces. By formulating the conflict resolution problem as a Multi-Agent Reinforcement Learning (RL) problem, we can train a policy offline to drive the vehicles towards their destinations safely and efficiently in a simplified discrete environment. During the online execution, each vehicle first simulates the interaction among vehicles with the trained policy to obtain its strategy, which is used to guide the computation of a reference trajectory. A distributed Model Predictive Controller (MPC) is then proposed to track the reference while avoiding collisions. The preliminary results show that the combination of RL and distributed MPC has the potential to guide vehicles to resolve conflicts safely and smoothly while being less computationally demanding than the centralized approach.
\end{abstract}

%%%%%%%%%%%%%%%%%%%%%%%%%%%%%%%%%%%%%%%%%%%%%%%%%%%%%%%%%%%%%%%%%%%%%%%%%%%%%%%%
\section{Introduction}
Autonomous vehicles (AVs) have the potential to revolutionize many aspects of our lives, including the way we commute and interact with the environment. But with this potential comes a unique set of challenges regarding how to resolve conflicts when vehicles encounter each other in highly constrained environments. Simple stop-and-go logic might lead to deadlocks since vehicles might have already blocked the paths of each other.

To address these challenges, researchers have developed various strategies for conflict resolution. Reinforcement Learning (RL)~\cite{wang_review_2022} is an approach that has been gaining attention in recent years. By learning from the environment and the interactions with other agents, autonomous vehicles can better understand the potential risks associated with each action and make decisions with higher long-term rewards. By formulating the conflict resolution problem as a Markov Decision Process (MDP), optimal actions can be obtained by either optimizing the expected cumulative reward in a level-$k$ game fashion~\cite{li_game_2018} or approximating the state-action value function with deep neural networks~\cite{li_optimizing_2019, yuan_deep_2022}. Despite being able to explore combinatorial actions of multiple agents, the existing RL approaches usually require approximations in agents' geometry or action space. These low-fidelity approximations would become insufficient in highly constrained spaces since vehicles often need to exploit their full motion capacities to maneuver around obstacles in close proximity.

Optimization-based methods are another approach to conflict resolution for autonomous vehicles. These methods incorporate analytical vehicle dynamics to optimize trajectories under constraints. A centralized Model Predictive Controller (MPC) was proposed in~\cite{riegger_centralized_2016} to optimize control actions for all permutations of crossing sequences. To reduce computation burden, a distributed MPC was designed with constraint prioritization~\cite{katriniok_distributed_2017}, and a decentralized controller was designed with the alternating direction method of multipliers (ADMM)~\cite{rey_fully_2018}. In highly constrained spaces, strong duality theorem can be applied to obtain an exact formulation of collision avoidance between two convex sets of arbitrary shape~\cite{zhang_optimization-based_2020}. This formulation was then used for a distributed MPC for multi-robot coordination~\cite{firoozi_distributed_2020}.

As discussed in~\cite{zhang_autonomous_2019}, good initial guesses are crucial for vehicles to find feasible trajectories to maneuver in highly constrained spaces. Our recently proposed method~\cite{shen_multi-vehicle_2022} used reinforcement learning in a discrete environment to search for configuration strategies, which guided a centralized model-based optimization problem to generate conflict-free trajectories under nonlinear, non-holonomic vehicle dynamics and exact collision avoidance constraints. However, the centralized optimization is computationally heavy and cannot adapt to uncertainties in online execution.

In this work, we provide a distributed algorithm based on the method proposed in~\cite{shen_multi-vehicle_2022}, but removing the need for a central trajectory planner. In particular,
\begin{enumerate}[label=(\roman*)]
    \item The conflict resolution problem is first addressed as a multi-agent RL problem in a discretized environment, where we can collect rollouts from all agents and train a shared policy offline. The reward function is designed to incentivize agents to reach their destinations quickly and penalize any possible collisions;
    \item When faced with a specific scenario, each vehicle uses a copy of the trained policy to simulate the interactions of all vehicles in the discrete environment. The results serve as a high-level strategy to guide the computation of a reference trajectory;
    \item During online execution, each vehicle follows its reference by solving a distributed MPC problem in real time. The collision avoidance constraints among vehicles and static obstacles are enforced for safety guarantees.
\end{enumerate}

This paper is organized as follows: Section~\ref{sec:formulation} formally defines the problem. Section~\ref{sec:learning} briefly describes our method as proposed in~\cite{shen_multi-vehicle_2022} to generate high-level strategies to resolve conflicts by deep RL. Section~\ref{sec:control} elaborates on the formulation of a distributed Model Predictive Controller. Section~\ref{sec:algo} presents the complete algorithm for distributed conflict resolution. The preliminary results of one example scenario are shown in Section~\ref{sec:results}. Finally, Section~\ref{sec:conclusion} concludes the paper.
\section{Problem Formulation}
\label{sec:formulation}

\subsection{Assumptions}
Considering $N_{\mathrm{v}}$ autonomous vehicles indexed by $i \in \mathcal{I} = \{0, \dots, N_{\mathrm{v}}-1\}$ in a highly constrained environment with $M$ static obstacles, $m \in \mathcal{M} = \{0, \dots, M-1\}$. The following assumptions are made:
\begin{enumerate}[label=(\roman*)]
    \item All vehicles have a map of the current environment and can localize themselves accurately;
    \item All vehicles are fully-autonomous and can communicate with each other to exchange information;
    \item All vehicles have identical body dimensions and dynamics.
\end{enumerate}

\subsection{Problem Statement}
\begin{figure}
\begin{center}
\includegraphics[width=\linewidth]{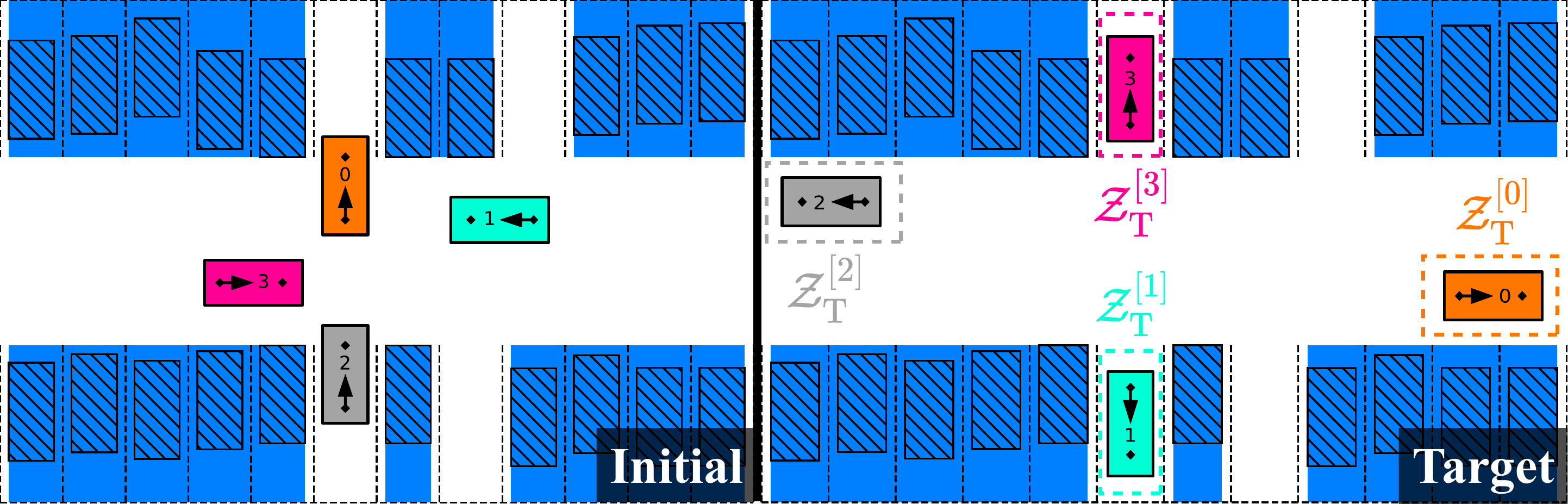}    % The printed column width is 8.4 cm.
\caption{An example scenario. The static obstacles are represented by blue rectangles, which are the over-approximation of the parked vehicles in diagonal hatches. Four vehicles 0-3 involved in this scenario are plotted in \textcolor{v0}{orange}, \textcolor{v1}{cyan}, \textcolor{v2}{grey}, and \textcolor{v3}{magenta}, while the short arrows point from the center of the rear axle to the center of the front axle, indicating the heading of the vehicles. Vehicles start from their initial states and aim to arrive at their target sets $\mathcal{Z}^{[i]}_{\mathrm{T}}, i \in \{0, 1, 2, 3\}$. }
\label{fig:scenario}
\end{center}
\end{figure}

The state $z$ of each vehicle is described by the position of its center of rear axle $(x, y)$, the heading angle $\psi$, the speed $v$, and the front steering angle $\delta_{f}$. In a conflict resolution scenario, each vehicle $i \in \mathcal{I}$ starts statically at 
% an initial pose $[x^{[i]}(0), y^{[i]}(0), \psi^{[i]}(0)]$ 
an initial state $z^{[i]}_{0}$
and aims to reach a target set $\mathcal{Z}^{[i]}_{\mathrm{T}}$ without collision against other vehicles or obstacles. An example scenario with four vehicles is illustrated in Fig.~\ref{fig:scenario}.

The geometry of each vehicle $i$ is a polytope computed by its dimension and real-time state $z^{[i]}$
\begin{equation}
    \mathbb{B}(z^{[i]}) := \{ p\in \mathbb{R}^2: G(z^{[i]}) p \leq g(z^{[i]})\}, \ i \in \mathcal{I},
    \label{eq:vehicle-polytope}
\end{equation}
and similarly, the static obstacles are described by polytopes
\begin{equation}
    \mathbb{O}^{[m]} := \{ p\in\mathbb{R}^2: A^{[m]} p \leq b^{[m]}\}, \ m \in \mathcal{M}.
    \label{eq:obs-polytope}
\end{equation}

Since vehicles operate at low speed, their dynamics follows the kinematic bicycle model:
\begin{equation}
    \dot{z} = f(z, u) := \left[
        \begin{matrix}
            \dot{x} \\ \dot{y} \\ \dot{\psi} \\ \dot{v} \\ \dot{\delta_f}
        \end{matrix}
    \right] = \left[
        \begin{matrix}
            v\cos(\psi) \\ v\sin(\psi) \\ \frac{v}{l_{\mathrm{wb}}}\tan(\delta_f) \\ a \\ \omega
        \end{matrix}
    \right], \ u = \left[
        \begin{matrix}
            a \\ \omega
        \end{matrix}
    \right],
    \label{eq:kin_model_ct}
\end{equation}
where the input $u$ consists of the acceleration $a$ and the steering rate $\omega$. Parameter $l_\mathrm{wb}$ represents the wheelbase. The discrete-time dynamics
\begin{equation}
    z_{k+1} = f_{\mathrm{dt}}(z_{k}, u_{k}, \tau) \label{eq:kin_model_dt}
\end{equation}
is obtained by discretizing~\eqref{eq:kin_model_ct} with the 4th-order Runge-Kutta method, where $\tau > 0$ is the samling time, and $z_{k}, u_{k}$ are the state and input at discrete time step $k$.

\subsection{Centralized Model Predictive Control}
Given a reference trajectory $\mathbf{z}_{\mathrm{ref}}^{[i]}$ that reaches the target set $\mathcal{Z}^{[i]}_{\mathrm{T}}$ from the initial state $z^{[i]}_{0}$ of each vehicle $i \in \mathcal{I}$, a centralized MPC problem can be formulated for control:
\begin{subequations}
\label{eq:centralized-formulation}
\begin{align}
    \min_{\mathbf{z}^{[i]}_{\cdot | t}, \mathbf{u}^{[i]}_{\cdot | t}} \ & \ \sum_{i \in \mathcal{I}} J^{[i]}\left(\mathbf{z}^{[i]}_{\cdot | t}, \mathbf{u}^{[i]}_{\cdot | t} \mid \mathbf{z}_{\mathrm{ref}}^{[i]}\right) \nonumber \\
    \text{s.t. } \ & z^{[i]}_{k+1 | t} = f_{\mathrm{dt}} \left(z^{[i]}_{k | t}, u^{[i]}_{k | t}, \tau\right), \\
    & z^{[i]}_{0 | t} = z^{[i]}(t) \\
    & z^{[i]}_{k | t} \in \mathcal{Z}, u^{[i]}_{k | t} \in \mathcal{U},\\
    & \mathrm{dist}\left(\mathbb{B}(z^{[i]}_{k | t}), \mathbb{O}^{(m)}\right)  \geq d_{\mathrm{min}}, \label{eq:collision-avoid-static}\\
    & \mathrm{dist}\left(\mathbb{B}(z^{[i]}_{k | t}), \mathbb{B}(z^{[j]}_{k | t})\right) \geq d_{\mathrm{min}},\label{eq:collision-avoid-vehicle}\\
    & \forall i, j \in \mathcal{I}, i \neq j, m \in \mathcal{M}, k \in {0, \dots, N} \nonumber,
\end{align}
\end{subequations}
where $\mathbf{z}^{[i]}_{\cdot | t} = \left\{z^{[i]}_{0 | t}, \dots, z^{[i]}_{N | t}\right\}$ and $\mathbf{u}^{[i]}_{\cdot | t} = \left\{u^{[i]}_{0 | t}, \dots, u^{[i]}_{N | t}\right\}$ denote the sequences of states and control inputs over the MPC look-ahead horizon $N$ for vehicle $i$. Vehicle states and inputs are constrained inside feasible sets $\mathcal{Z}$ and $\mathcal{U}$. Constraints~\eqref{eq:collision-avoid-static} and~\eqref{eq:collision-avoid-vehicle} enforce the collision avoidance between static obstacles and other vehicles., where $\mathrm{dist}(\cdot, \cdot) \in \mathbb{R}$ measures the distance between to polytopes, and $d_{\mathrm{min}}$ is a safety threshold.

Several challenges impede our ability to apply this centralized approach effectively:
\begin{enumerate}[label=(\roman*)]
    \item It is hard to find such reference trajectories $\mathbf{z}_{\mathrm{ref}}^{[i]}$ that while following them, the vehicles can easily find smooth maneuvers to avoid collisions and stay feasible;
    \item Constraints~\eqref{eq:collision-avoid-static} and~\eqref{eq:collision-avoid-vehicle} make \eqref{eq:centralized-formulation} a bi-level optimization problem, therefore intractable to solve with the off-the-shelf solvers;
    \item It is computationally heavy for a central commander to solve problem~\eqref{eq:centralized-formulation} iteratively due to the pairwise collision avoidance between all vehicles and all obstacles, therefore not amendable for online deployment.
\end{enumerate}

In the remainder of this paper, we will elaborate on our method to compute high-quality reference trajectories based on RL-generated strategies and the distributed reformulation of~\eqref{eq:centralized-formulation} so that it becomes real-time capable.

\section{Generating Strategies with RL}
\label{sec:learning}
This section describes our method to generate strategies for vehicles to compute their reference trajectories. The strategies are represented as sequences of tactical vehicle configurations and are computed by constructing the conflict resolution problem as a discrete multi-agent reinforcement learning problem. A part of this section is extracted from~\cite{shen_multi-vehicle_2022} and reported here just for the sake of completeness and readability. The reader is referred to~\cite{shen_multi-vehicle_2022} for an in-depth discussion on strategy generation.

\subsection{State and Observation Space}

\begin{figure}
	\centering
	\begin{subfigure}[t]{0.59\columnwidth}
		\centering
		\includegraphics[width=\textwidth]{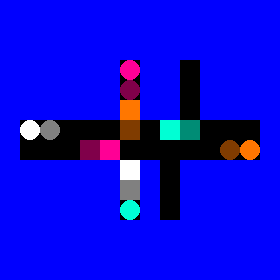}
	    \caption{Discrete grid environment}
	    \label{fig:discrete-env}
	\end{subfigure}%
	~
	\begin{subfigure}[t]{0.39\columnwidth}
		\centering
		\includegraphics[width=\textwidth]{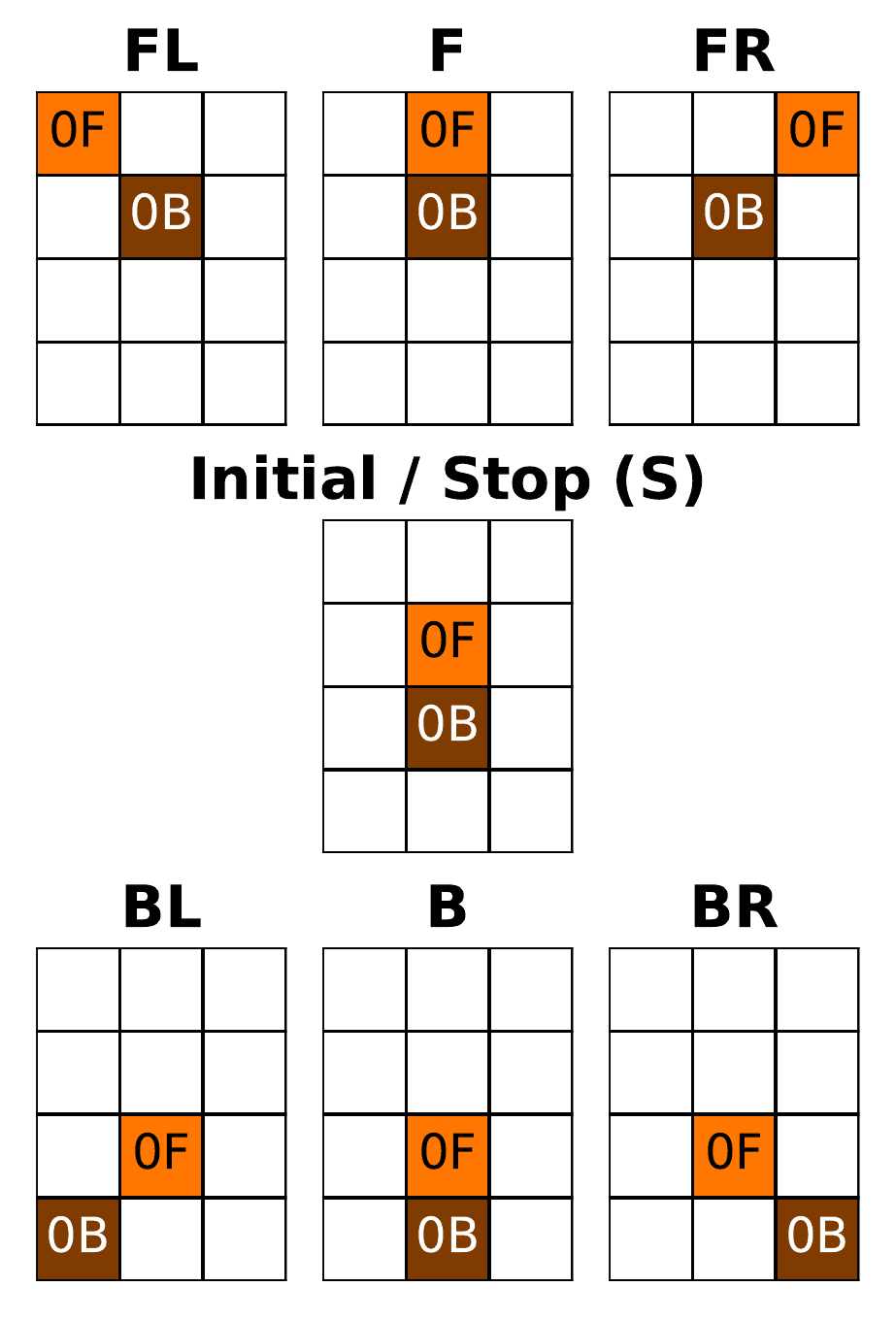}
		\caption{Dynamics in grids}
		\label{fig:grid-dynamics}
	\end{subfigure}
	\caption{Vehicles in the discrete environment. (a) Vehicle bodies are shown as squares, and their destinations are shown as circles. \textcolor{v0}{Orange}, \textcolor{v1}{cyan}, \textcolor{v2}{white/grey}, and \textcolor{v3}{magenta} represent different vehicles, and the brightness change distinguishes their front and back sides. Obstacles are shown in blue and free spaces are in black. (b) Single-vehicle dynamics in a free grid map. ``0F'' and ``0B'' mark the front and back of vehicle $0$.}
\end{figure}

In the discrete RL problem, each vehicle's state and destination are represented by the squares of its ``front'' and ``back'' sides in a discrete grid environment. Fig.~\ref{fig:discrete-env} shows the discrete representation corresponding to the initial conditions and target sets of the scenario example in Fig.~\ref{fig:scenario}. Furthermore, RGB images like Fig.~\ref{fig:discrete-env} are taken as the agents' observations of the current step.

\subsection{Action Space}
\label{sec:action-space}
Each vehicle can take $7$ discrete actions in the grid map: \{Stop (S), Forward (F), Forward Left (FL), Forward Right (FR), Backward (B), Backward Left (BL), Backward Right (BR)\}. The corresponding state transitions of a single vehicle in a free grid map are demonstrated in Fig.~\ref{fig:grid-dynamics}. When a vehicle collides with any obstacle or vehicle, it will be ``bounced back'' so that its state remains unchanged.

\subsection{Reward}
The reward is designed to penalize collision, the ``Stop (S)'' action, the distance away from the vehicle's destination, and time consumption. The vehicle will also receive a huge positive reward upon arriving at its destination.

\subsection{Policy Learning}
By collecting rollouts from all vehicles in extensive simulations, we train a CNN-based deep Q-network with parameter $\theta$ to approximate the observation-action value function $Q(o, a|\theta)$, where $o$ is an RGB image like Fig.~\ref{fig:discrete-env} and $a$ is among the actions introduced in Sec.~\ref{sec:action-space}. The policy is expressed as $\pi_{\theta} = \mathrm{argmax}_a Q(o, a | \theta)$.

\subsection{Strategy}

\begin{figure}
\begin{center}
\includegraphics[width=\linewidth]{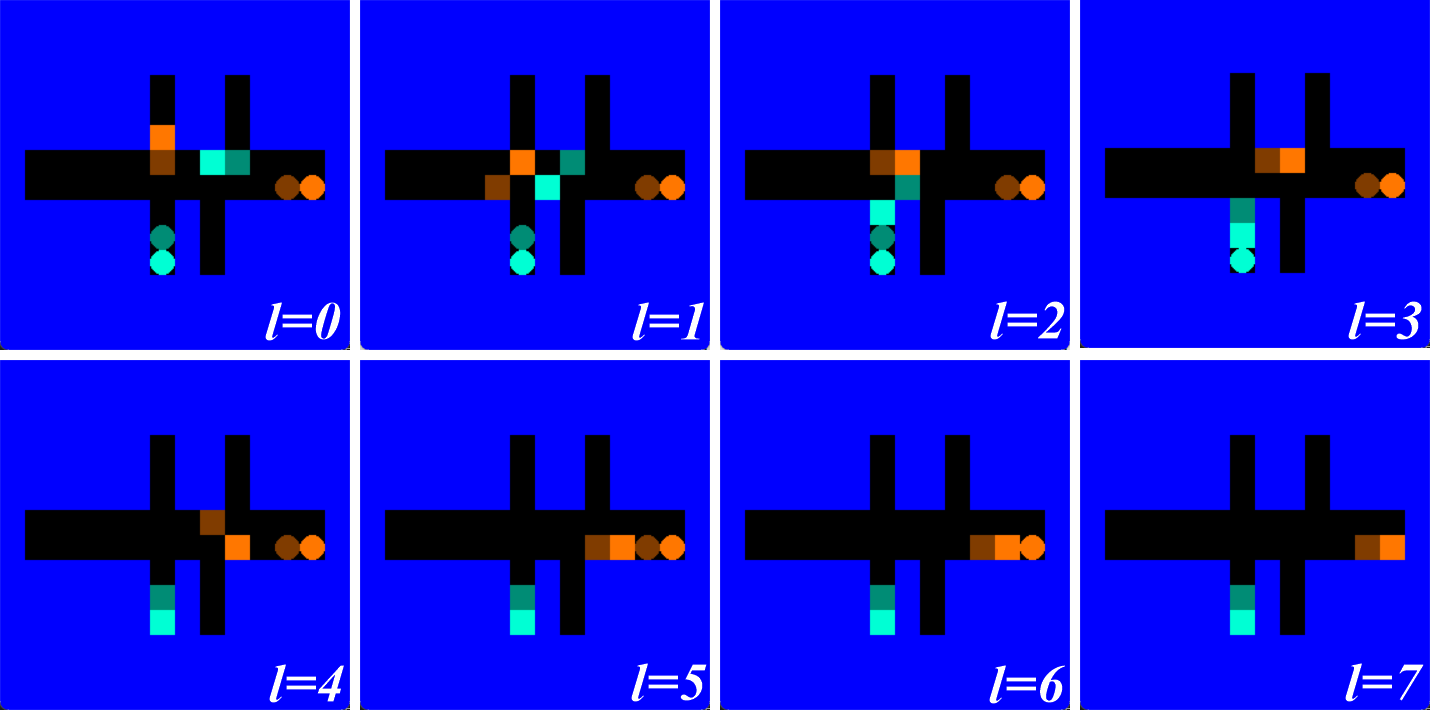}    % The printed column width is 8.4 cm.
\caption{The strategies of two vehicles to resolve conflicts}
\label{fig:strategy-2v}
\end{center}
\end{figure}

Given a specific conflict scenario, each vehicle can use the trained policy $\pi_{\theta^{*}}$ to simulate the interactions of all vehicles in the discrete environment until the conflict is resolved. A two-vehicle example is shown in Fig.~\ref{fig:strategy-2v}. We use ``strategy'' to refer to the sequence of vehicle configurations reflected by the squares in the discrete grid environment. By applying transformations, the squares in the grid map become strategy-guided sets in ground coordinates
\begin{equation*}
    \mathbf{Z}^{[i]} = \left\{\Bar{\mathcal{Z}}_{\mathrm{F}, l}^{[i]}, \Bar{\mathcal{Z}}_{\mathrm{B}, l}^{[i]}\right\}_{l=0}^{L^{[i]}-1}
\end{equation*}
where $L^{[i]}$ is the number of steps that
vehicle $i$ takes to reach its destination, $\Bar{\mathcal{Z}}_{\mathrm{F}, l}^{[i]} \subset \mathbb{R}^{2 \times 2}$ and $\Bar{\mathcal{Z}}_{\mathrm{B}, l}^{[i]} \subset \mathbb{R}^{2 \times 2}$ are convex sets for the front and back of the vehicle $i$ at step $l$. 

Denote by $T_{\mathrm{s}}$ the time period between two strategy steps $l$ and $l+1$, the vehicle configurations in the continuous space are guided such that at time $T_{\mathrm{s}}l$, the center of vehicle rear axle $(x^{[i]}, y^{[i]})$ is inside the set $\Bar{\mathcal{Z}}_{\mathrm{B}, l}^{[i]}$, and the center of vehicle front axle $(x_{\mathrm{F}}^{[i]}, y_{\mathrm{F}}^{[i]})$ is inside the set $\Bar{\mathcal{Z}}_{\mathrm{F}, l}^{[i]}$, formally
\begin{subequations}
    \label{eq:strategy-guided-config-constraints}
    \begin{align}
        \left[ x^{[i]}(T_{\mathrm{s}}l), y^{[i]}(T_{\mathrm{s}}l)\right]^{\top} & \in \Bar{\mathcal{Z}}_{\mathrm{B}, l}^{[i]} \\
        \left[ x_{\mathrm{F}}^{[i]}(T_{\mathrm{s}}l), y_{\mathrm{F}}^{[i]}(T_{\mathrm{s}}l)\right]^{\top} & \in \Bar{\mathcal{Z}}_{\mathrm{F}, l}^{[i]}, \\
        x_{\mathrm{F}}^{[i]}(T_{\mathrm{s}}l) = x^{[i]}(T_{\mathrm{s}}l) + l_{\mathrm{wb}} & \cos\left(\psi^{[i]}(T_{\mathrm{s}}l)\right) \\
        y_{\mathrm{F}}^{[i]}(T_{\mathrm{s}}l) = y^{[i]}(T_{\mathrm{s}}l) + l_{\mathrm{wb}} & \sin\left(\psi^{[i]}(T_{\mathrm{s}}l)\right)\\
        \forall l = 0, \dots, L^{[i]} -1, i &\in \mathcal{I}, \nonumber
    \end{align}
\end{subequations}

\section{Strategy-guided Distributed MPC}
\label{sec:control}
In this section, we describe our approach to enable each vehicle $i$ to compute its reference trajectory $\mathbf{z}_{\mathrm{ref}}^{[i]}$ by reformulating collision avoidance (CA) constraints and leveraging the strategy-guided constraints~\eqref{eq:strategy-guided-config-constraints}. Then, a distributed MPC is proposed for each vehicle $i$ to solve independently for its control input in real-time.

\subsection{Collision Avoidance (CA) Against Static Obstacles}
According to~\cite{zhang_optimization-based_2020, firoozi_distributed_2020}, the CA constraint~\eqref{eq:collision-avoid-static} between the pair of vehicle $i \in \mathcal{I}$ and static obstacle $m \in \mathcal{M}$ can be reformulated as a feasibility problem:
\begin{subequations}
\label{eq:obca-obs}
\begin{align}
    \exists \lambda^{[i]}_{m} \geq 0, \mu^{[i]}_{m} \geq 0, s^{[i,m]} \ : \|s^{[i,m]}\| & \leq 1, \\
    - g(z^{[i]})^{\top}\lambda^{[i]}_{m} - b^{[m], \top} \mu^{[i]}_{m} & \geq d_{\mathrm{min}}, \\
    G(z^{[i]}) \lambda^{[i]}_{m} + s^{[i,m]} & = 0, \\
    A^{[m]} \mu^{[i]}_{m} - s^{[i,m]} & = 0,
\end{align}
\end{subequations}
where $G(\cdot), g(\cdot), A^{[m]}, b^{[m]}$ describe polytopes $\mathbb{B}(z^{[i]}), \mathbb{O}^{[m]}$ as introduced in~\eqref{eq:vehicle-polytope}, \eqref{eq:obs-polytope}.

\subsection{Optimization-based Reference Computation}
By incorporating the strategy-guided constraints~\eqref{eq:strategy-guided-config-constraints} and the collision avoidance constraints~\eqref{eq:obca-obs}, we can compute the reference trajectory $\mathbf{z}_{\mathrm{ref}}^{[i]}$ of each vehicle $i$ by solving the optimal control problem
\begin{subequations}
\label{eq:reference-computation}
\begin{align}
    \min_{\substack{\mathbf{z}^{[i]}, \mathbf{u}^{[i]} \\ \boldsymbol{\lambda}^{[i]}, \boldsymbol{\mu}^{[i]}, \mathbf{s}^{[i, \cdot]}}} \ & \int_{t=0}^{T^{[i]}_{\mathrm{s}}L^{[i]}} c\left(z^{[i]}(t), u^{[i]}(t)\right) dt\nonumber \\
    \text{s.t. } \ & \dot{z}^{[i]}(t) = f(z^{[i]}(t), u^{[i]}(t)), \label{eq:ref-comp-dynamics}\\
    & z^{[i]}(t) \in \mathcal{Z}, u^{[i]}(t) \in \mathcal{U}, \label{eq:ref-comp-state-input-limit}\\
    & z^{[i]}(0) = z^{[i]}_0, \label{eq:ref-comp-init}\\
    & z^{[i]}(T_{\mathrm{s}}L^{[i]}) \in \mathcal{Z}_{\mathrm{T}}^{[i]},\label{eq:ref-comp-final}\\
    & \text{Strategy-guided Constraints } \eqref{eq:strategy-guided-config-constraints} \nonumber,\\
    & \text{CA~\eqref{eq:obca-obs} against obstacle } m \nonumber,\\
    & \forall 0 \leq t \leq T_{\mathrm{s}}L^{[i]}, m \in \mathcal{M}. \nonumber
\end{align}
\end{subequations}
Intuitively, the reference trajectory $\mathbf{z}_{\mathrm{ref}}^{[i]}$ is computed such that
\begin{enumerate}[label=(\roman*)]
    \item it is kinematically feasible by obeying the vehicle dynamics constraint~\eqref{eq:ref-comp-dynamics} and state and input limits~\eqref{eq:ref-comp-state-input-limit};
    \item it starts from the initial state $z_{0}^{[i]}$~\eqref{eq:ref-comp-init} and reaches the target set $\mathcal{Z}_{\mathrm{T}}^{[i]}$ in the end~\eqref{eq:ref-comp-final};
    \item it follows the strategy-guided configurations defined by $\mathbf{Z}^{[i]}$ in a sequential manner, at a time period of $T_{\mathrm{s}}$;
    \item it avoids all static obstacles in the environment. Note that we don't enforce the inter-vehicle collision avoidance constraints for reference computation since other vehicles' states are unavailable.
\end{enumerate}
Despite the continuous time formulation, problem~\eqref{eq:reference-computation} can be discretized with orthogonal collocation on finite elements~\cite{biegler_nonlinear_2010}, where the number of intervals is given by the number of strategy steps $L^{[i]}$, and the length of each interval is $T_{\mathrm{s}}$. The interpolation polynomials for collocation are also used to sample reference vehicle states during online control.

\subsection{Collision Avoidance (CA) Against Other Vehicles}
Similar to~\eqref{eq:obca-obs}, the CA constraint~\eqref{eq:collision-avoid-vehicle} between the vehicle pair $i, j \in \mathcal{I}, i \neq j$ is reformulated as
\begin{subequations}
\label{eq:obca-vehicle}
\begin{align}
    \exists \lambda^{[i]}_{j} \geq 0, \mu^{[i]}_{j} \geq 0, s^{[i,j]} \ : \|s^{[i,j]}\| & \leq 1, \\
    - g(z^{[i]})^{\top}\lambda^{[i]}_{j} - g(z^{[j]})^{\top} \mu^{[i]}_{j} & \geq d_{\mathrm{min}}, \\
    G(z^{[i]}) \lambda^{[i]}_{j} + s^{[i,j]} & = 0, \\
    G(z^{[j]}) \mu^{[i]}_{j} - s^{[i,j]} & = 0.
\end{align}
\end{subequations}

\subsection{Distributed MPC}
At time $t$, each vehicle $i$ independently computes its own state and input trajectory $\mathbf{z}_{\cdot \mid t}^{[i]}, \mathbf{u}_{\cdot \mid t}^{[i]}$ over the horizon $N$ by solving the following optimization problem
\begin{subequations}
\label{eq:distributed-formulation}
\begin{align}
    \min_{ \substack{\mathbf{z}^{[i]}_{\cdot | t}, \mathbf{u}^{[i]}_{\cdot | t} \\ \boldsymbol{\lambda}^{[i]}_{\cdot | t}, \boldsymbol{\mu}^{[i]}_{\cdot | t}, \mathbf{s}^{[i, \cdot]}_{\cdot | t}} } \ & \ J^{[i]}\left(\mathbf{z}^{[i]}_{\cdot | t}, \mathbf{u}^{[i]}_{\cdot | t} \mid \mathbf{z}_{\mathrm{ref}}^{[i]}\right) \nonumber \\
    \text{s.t. } \ & z^{[i]}_{k+1 | t} = f_{\mathrm{dt}} \left(z^{[i]}_{k | t}, u^{[i]}_{k | t}, \tau\right), \\
    & z^{[i]}_{0 | t} = z^{[i]}(t) \\
    & z^{[i]}_{k | t} \in \mathcal{Z}, u^{[i]}_{k | t} \in \mathcal{U},\\
    & \text{CA~\eqref{eq:obca-obs} against obstacle } m  \nonumber\\
    & \text{CA~\eqref{eq:obca-vehicle} against vehicle $j$, given $\Bar{\mathbf{z}}^{[j]}_{\cdot \mid t}$}   \nonumber\\
    & \forall j \in \mathcal{I} \backslash \{i\}, m \in \mathcal{M}, k \in {0, \dots, N} \nonumber,
\end{align}
\end{subequations}
where $\Bar{\mathbf{z}}^{[j]}_{\cdot \mid t}$ denotes the known look-ahead prediction from another vehicle $j$, which is obtained by inter-vehicle communication. The distributed formulation above addresses the challenges of centralized formulation~\eqref{eq:centralized-formulation} by:
\begin{enumerate}[label=(\roman*)]
    \item following a ``high-quality'' strategy-guided reference trajectory $\mathbf{z}_{\mathrm{ref}}^{[i]}$ which already encodes the required maneuver to resolve conflict through the guidance of strategy;
    \item using smooth CA constraints such that the problem is solvable by gradient- or Hessian-based solvers;
    \item solving the decision variables for vehicle $i$ only so it can be real-time capable. Furthermore, the complexity of the problem only grows linearly with the number of obstacles and other vehicles. 
\end{enumerate}
\section{Distributed Conflict Resolution Algorithm}
\label{sec:algo}

This section presents the Algorithm~\ref{algo:algo} for distributed conflict resolution in highly constrained space. 

Line~\ref{algo:construct-map} to~\ref{algo:record-strategy} correspond to the process described in Section~\ref{sec:learning}. After getting the initial states $z^{[i]}_0$ and target sets $\mathcal{Z}^{[i]}_{\mathrm{T}}$ of all vehicles, each vehicle $i$ sets up a discrete grid environment as Fig.~\ref{fig:discrete-env} to simulate the interactions among all vehicles until it reaches its destination. Note that for each scenario, all vehicles create identical discrete environments and implement the same policy $\pi_{\theta^{*}}$ for all vehicles. As a result, the outcomes obtained by one vehicle will be identical to those obtained by any other vehicle. The sequence of discrete steps performed by vehicle $i$ itself is recorded as a strategy to generate sets $\mathbf{Z}^{[i]}$ to guide the vehicle configurations.
% Since for any given scenario, all vehicles construct identical discrete environments and share the same policy $\pi_{\theta^{*}}$, the interactions simulated by one vehicle are identical to that by others. 
% no matter which vehicle the conflict is resolved in a deterministic way, despite being computed by each vehicle independently.

After the reference trajectory $\mathbf{z}^{[i]}_{\mathrm{ref}}$ is computed from~\eqref{eq:reference-computation}, each vehicle $i$ will firstly interpolate $N$ states from $\mathbf{z}^{[i]}_{\mathrm{ref}}$ along its MPC horizon as its initial look-ahead prediction $\mathbf{z}_{\cdot \mid -1}^{[i]}$ and broadcast it to all other vehicles, such that all vehicles can solve problem~\eqref{eq:distributed-formulation} starting from $t=0$. 

At time $t$, the latest information about another vehicle $j$ is $\mathbf{z}_{\cdot \mid t-1}^{[j]} = \left\{z^{[j]}_{0 | t-1}, \dots, z^{[j]}_{N | t-1}\right\}$ received from time $t-1$. Therefore, in line~\ref{algo:shift}, we shift $\mathbf{z}_{\cdot \mid t-1}^{[j]}$ one step ahead such that $\Bar{\mathbf{z}}_{\cdot \mid t}^{[j]} = \left\{z^{[j]}_{1 | t-1}, \dots, z^{[j]}_{N | t-1}, z^{[j]}_{N | t-1}\right\}$. Additionally, since the solution of the non-convex, nonlinear programming problem~\eqref{eq:distributed-formulation} critically depends on the initial guesses, we shift the optimal solutions $\mathbf{z}_{\cdot \mid t-1}^{[i]}, \mathbf{u}_{\cdot \mid t-1}^{[i]}, \boldsymbol{\lambda}^{[i]}_{\cdot | t-1}, \boldsymbol{\mu}^{[i]}_{\cdot | t-1}, \mathbf{s}^{[i, \cdot]}_{\cdot | t-1}$ from time $t-1$ to warm start the solver at time $t$.

\begin{algorithm}
\caption{Distributed Conflict Resolution}
\label{algo:algo}
\begin{algorithmic}[1]
    \Require initial states $z^{[i]}_0$ and target sets $\mathcal{Z}^{[i]}_{\mathrm{T}}$ of all vehicles $i \in \mathcal{I}$, static obstacles, trained RL policy $\pi_{\theta^{*}}$
    \For{all vehicle $i \in \mathcal{I}$ in parallel}
        \State Construct a grid map with $z^{[i]}_0, \forall i \in \mathcal{I}$ and obstacles\label{algo:construct-map}
        \While{vehicle $i$ has not reach its destination}
            \State Simulate vehicle interactions with $\pi_{\theta^{*}}$\label{algo:simulate}
            \State Record each step as the strategy of vehicle $i$\label{algo:record-strategy}
        \EndWhile
        \State Compute sets $\mathbf{Z}^{[i]}$ based on the recorded strategy
        \State Compute reference trajectory $\mathbf{z}_{\mathrm{ref}}^{[i]}$ by solving~\eqref{eq:reference-computation}
        \State Compute the initial $\mathbf{z}_{\cdot \mid -1}^{[i]}$ by sampling along $\mathbf{z}_{\mathrm{ref}}^{[i]}$
        \State Broadcast $\mathbf{z}_{\cdot \mid -1}^{[i]}$ to other vehicles $j \in \mathcal{I} \backslash \{i\}$
        \State Initialize $\boldsymbol{\lambda}^{[i]}_{\cdot | -1}, \boldsymbol{\mu}^{[i]}_{\cdot | -1}, \mathbf{s}^{[i, \cdot]}_{\cdot | -1}$ for CA
    \EndFor
    \For{t = 0, 1, $\dots$, $\infty$}\label{algo:for-loop}
        \For{all vehicle $i \in \mathcal{I}$ in parallel}
            \State Shift the latest $\mathbf{z}_{\cdot \mid t-1}^{[j]}$ to get $\Bar{\mathbf{z}}_{\cdot \mid t}^{[j]}$, for all $j \in \mathcal{I} \backslash \{i\}$\label{algo:shift}
            \State Compute $\mathbf{z}_{\cdot \mid t}^{[i]}, \mathbf{u}_{\cdot \mid t}^{[i]}, \boldsymbol{\lambda}^{[i]}_{\cdot | t}, \boldsymbol{\mu}^{[i]}_{\cdot | t}, \mathbf{s}^{[i, \cdot]}_{\cdot | t}$ by solving~\eqref{eq:distributed-formulation}
            \State Broadcast $\mathbf{z}_{\cdot \mid t}^{[i]}$ to other vehicles $j \in \mathcal{I} \backslash \{i\}$
            \State Apply $u_{0 \mid t}^{[i]}$ to move forward\label{algo:apply-control}
        \EndFor
    \EndFor
\end{algorithmic} 
\end{algorithm}
 
\section{Results}
\label{sec:results}

The following section presents the details of a conflict scenario, as depicted in Fig.~\ref{fig:scenario}, and the simulation results of the proposed distributed Algorithm~\ref{algo:algo}. To maximize clarity, we have selected the scenario in Fig.~\ref{fig:scenario} as it is the most complex among all scenarios tested. For the source code and results of other scenarios, please refer to: https://bit.ly/rl-cr.

\begin{table}[t]
\centering
\caption{Constraints for Initial and Final Poses}
\label{tab:initial-final-pose}
\begin{tabular}{ccccc}
\toprule
$i$ & Initial Pose $(x,y,\psi)$ & Final $x$ & Final $y$ & Final $\psi$ \\ \midrule
0 & $(16.25, 18.75, \frac{1}{2}\pi)$  & $[27.5, 30]$ & $[15, 17.5]$ & $0$ \\
\midrule
1 & $(23.75, 18.75, \pi)$  & $[15, 17.5]$ & $[10, 12.5]$ & $-\frac{1}{2}\pi$ \\       
\midrule
2 & $(16.25, 11.25, \frac{1}{2}\pi)$  & $[5, 7.5]$ & $[17.5, 20]$ & $\pi$ \\   
\midrule
3 & $(11.25, 16.25, 0)$  & $[15, 17.5]$ & $[22.5, 25]$ & $\frac{1}{2}\pi$
\end{tabular}
\end{table}

The entire parking lot region is of size 30m $\times$ 20m where most spots are occupied, creating a highly constrained environment. The vehicle poses in their initial states $z_{0}^{[i]}$ and target sets $\mathcal{Z}_{\mathrm{T}}^{[i]}$ are described in Table~\ref{tab:initial-final-pose}. All other state and input components are 0 at the initial time step. The vehicle bodies are 3.9m $\times$ 1.8m rectangles, and their wheelbases are 2.5m. The states and inputs of vehicles are under operation limits that $v^{[i]} \in [-2.5, 2.5] \mathrm{m/s}$, $\delta_{f}^{[i]} \in [-0.85, 0.85] \mathrm{rad}$, $a^{[i]} \in [-1.5, 1.5] \mathrm{m/s^2}$, $w^{[i]} \in [-1, 1] \mathrm{rad/s}$. Throughout this work, we set the safety threshold as $d_{\mathrm{min}} = 0.05$m. The optimization problems are coded with CasADi~\cite{andersson_casadi_2019} and solved by IPOPT~\cite{wachter_implementation_2006} with the linear solver HSL\_MA97~\cite{rees_hsl_2022}.

\subsection{Strategy-guided Reference Trajectories}

\begin{figure}
\begin{center}
\includegraphics[width=0.97\linewidth]{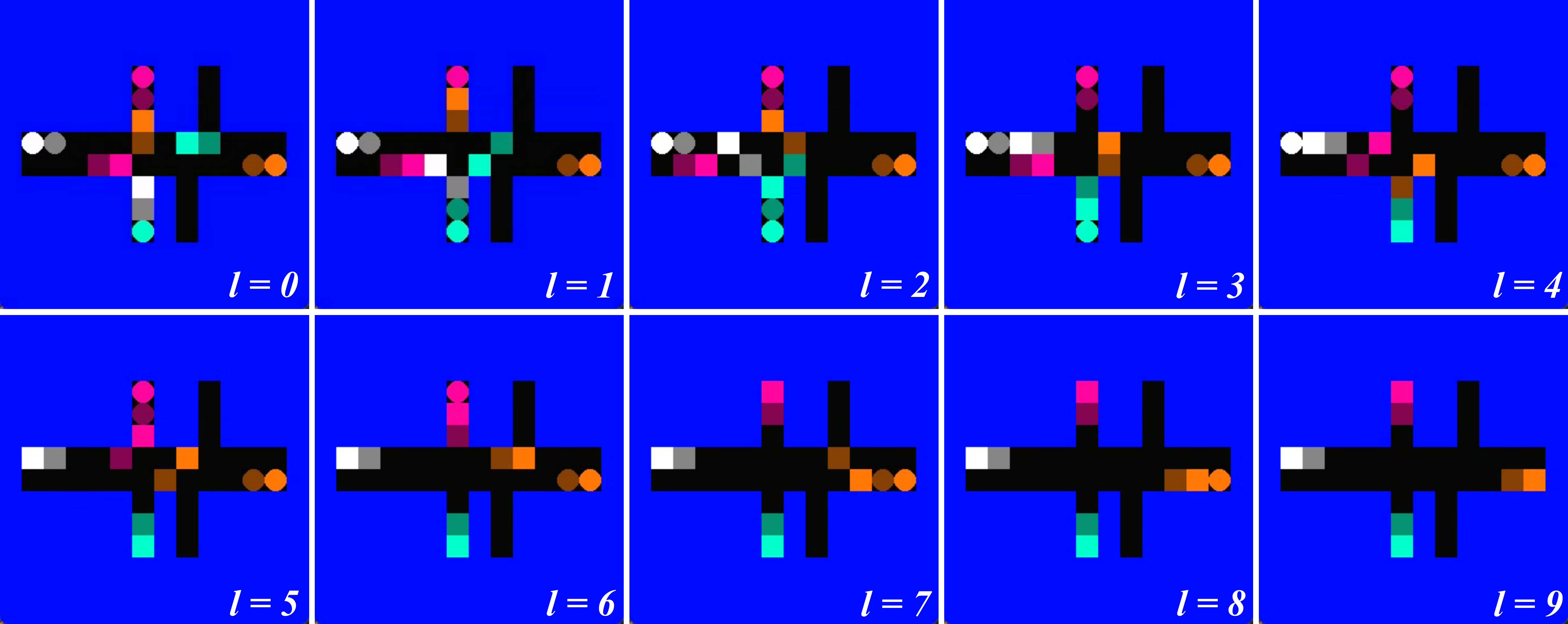}    % The printed column width is 8.4 cm.
\caption{Steps of each vehicle in the discrete environment to resolve conflict.}
\label{fig:rl-results}
\end{center}
\end{figure}

By using the RL policy $\pi_{\theta^{*}}$ to simulate vehicle interaction in the discrete environment, each vehicle obtains a sequence of discrete steps as its strategy to resolve conflict, as shown in Fig.~\ref{fig:rl-results}. It can be seen from the figure that \textcolor{v0}{vehicle 0 (orange)} firstly drives forward into the upper spot to make spaces for other vehicles, then backs up and changes its heading angle by utilizing the space in the bottom spot; \textcolor{v1}{vehicle 1 (cyan)} and \textcolor{v2}{vehicle 2 (grey)} immediately drive towards their destinations; \textcolor{v3}{vehicle 3 (magenta)} firstly backs up to avoid collisions, then drives towards the upper spot. The number of strategy steps $L^{[i]}$ required by each vehicle are $L^{[0]} = 10, L^{[1]}=5, L^{[2]}=6, L^{[3]}=8$.

\begin{figure*}
\begin{center}
\includegraphics[width=\linewidth]{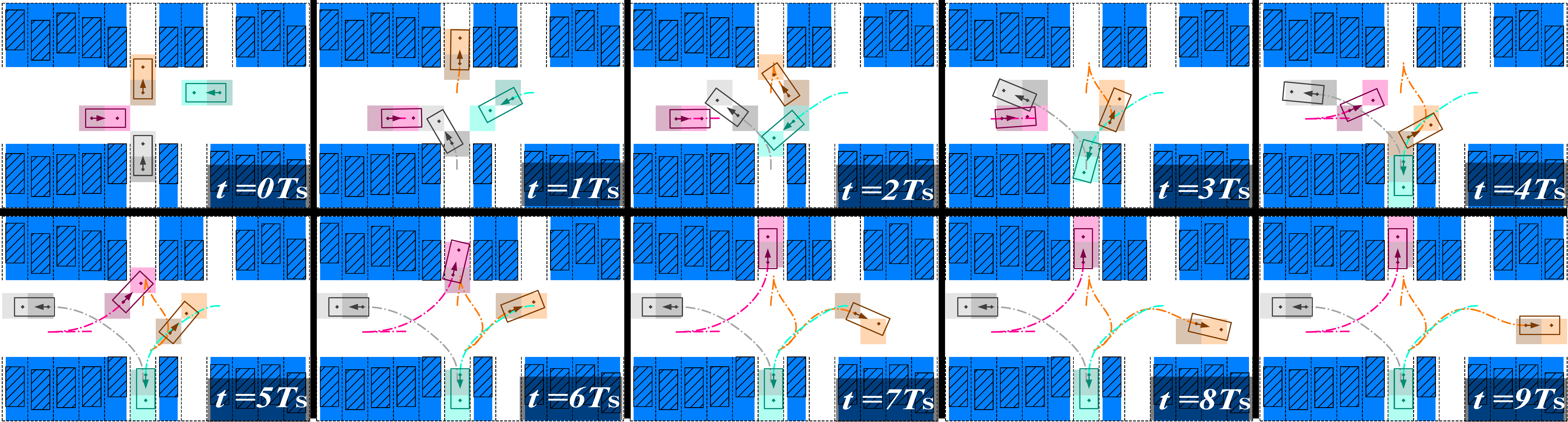}    % The printed column width is 8.4 cm.
\caption{Strategy-guided sets $\mathbf{Z}^{[i]}$ at time $t=lT_{\mathrm{s}}, l = 0, \dots, 9$ and the resulting vehicle reference trajectories. The sets, vehicle bodies, and trajectories are plotted in \textcolor{v0}{orange}, \textcolor{v1}{cyan}, \textcolor{v2}{grey}, and \textcolor{v3}{magenta} to represent \textcolor{v0}{vehicle 0}, \textcolor{v1}{vehicle 1}, \textcolor{v2}{vehicle 2} and \textcolor{v3}{vehicle 3} respectively.}
\label{fig:4v-ref-poses}
\end{center}
\end{figure*}

Corresponding to the strategies in Fig.~\ref{fig:rl-results}, the strategy-guided sets $\mathbf{Z}^{[i]}$ for different vehicles are computed and shown as the \textcolor{v0}{orange}, \textcolor{v1}{cyan}, \textcolor{v2}{grey}, \textcolor{v3}{magenta} squares in Fig.~\ref{fig:4v-ref-poses}. We use the stage cost $c(z, u) = \psi^2 + v^2 w^2 + a^2 + 1$ in problem~\eqref{eq:reference-computation} to describe the passenger comfort, the amount of actuation, and the time consumption. The time period between the two strategy steps is $T_{\mathrm{s}} = 3$s. We use 5-th order Lagrange interpolation polynomial and Gauss-Radau roots for collocation. By solving problem~\eqref{eq:reference-computation}, we obtain the reference trajectories $\mathbf{z}_{\mathrm{ref}}^{[i]}$ for all vehicle $i \in \mathcal{I}$ as plotted in Fig.~\ref{fig:4v-ref-poses}. The vehicle configurations at $t=l T_{\mathrm{s}}, l=0, \dots, 9$ are also drawn to demonstrate the effect of strategy-guided constraints~\eqref{eq:strategy-guided-config-constraints}. Note that since we only enforce collision avoidance constraints~\eqref{eq:obca-obs} in~\eqref{eq:reference-computation}, the reference trajectories are not guaranteed to be collision-free against each other, as reflected by vehicle configurations at time $t=3T_{\mathrm{s}}$ in Fig.~\ref{fig:4v-ref-poses}.

\subsection{Distributed Online Control}

\begin{figure}
\begin{center}
\includegraphics[width=\linewidth]{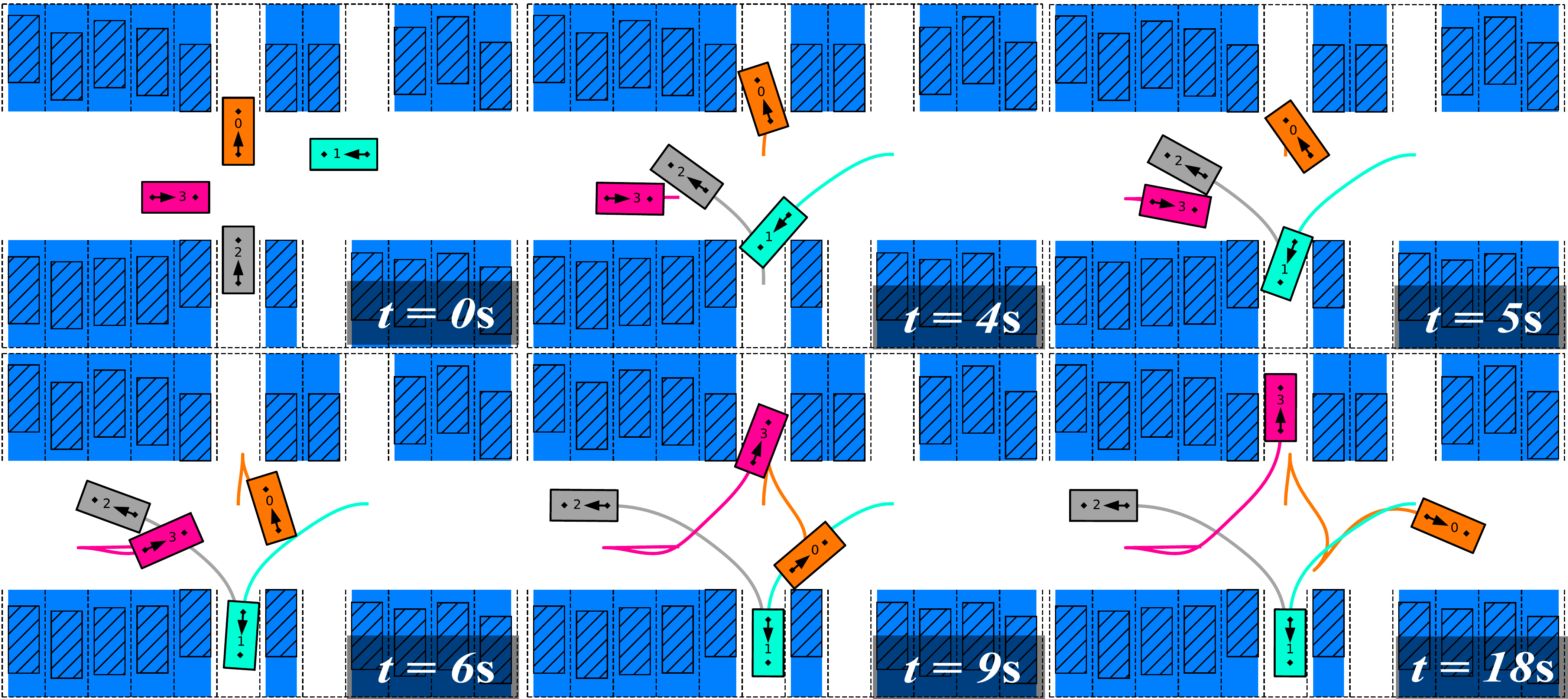}    % The printed column width is 8.4 cm.
\caption{Snapshots during online control}
\label{fig:final-poses}
\end{center}
\end{figure}

\begin{figure}
\begin{center}
\includegraphics[width=\linewidth]{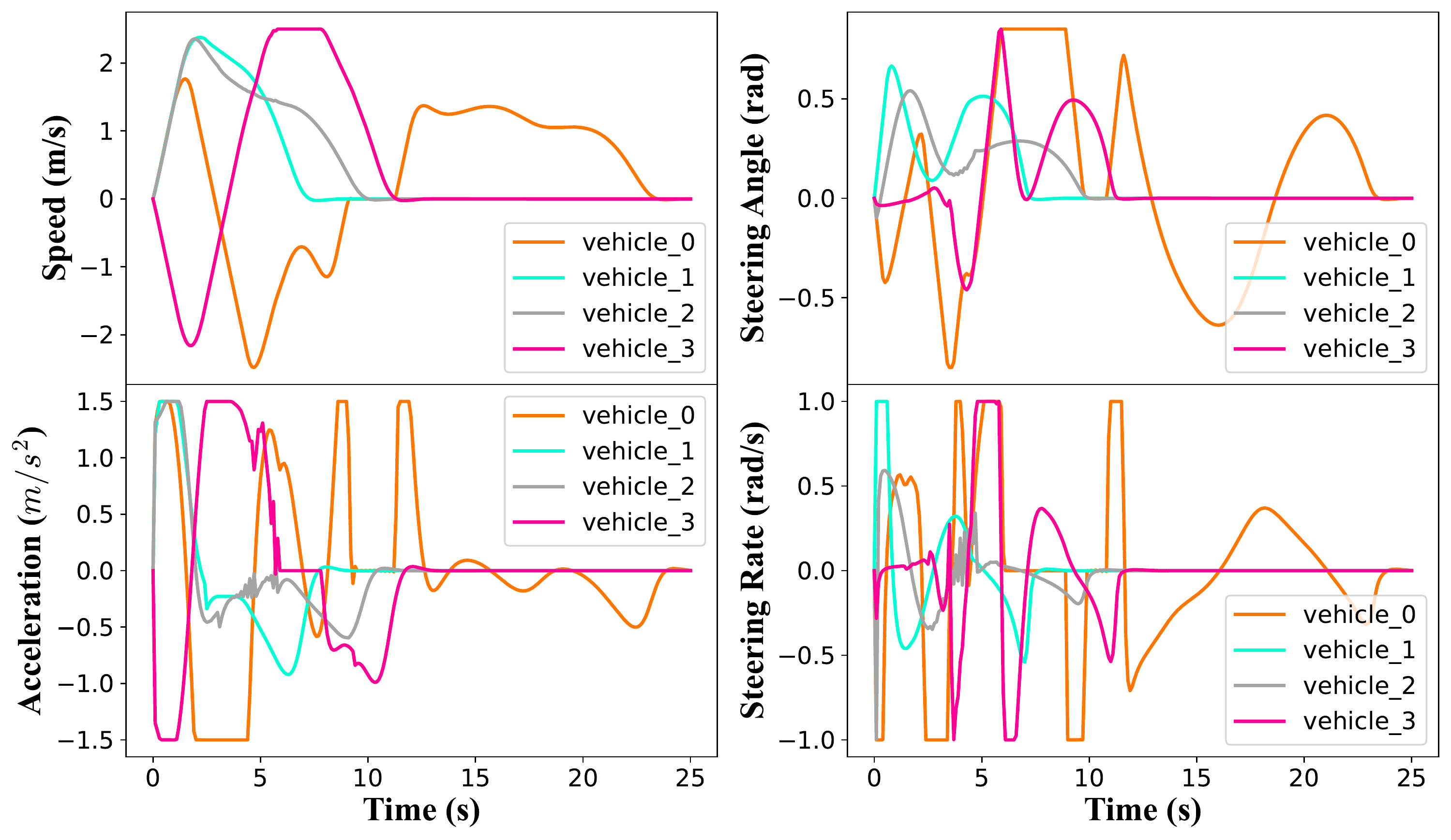}    % The printed column width is 8.4 cm.
\caption{Profile of vehicle speed, front steering angle, acceleration, and steering rate}
\label{fig:state-input-profile}
\end{center}
\end{figure}

\begin{figure}
	\centering
	\begin{subfigure}[t]{0.64\columnwidth}
		\centering
		\includegraphics[width=\textwidth]{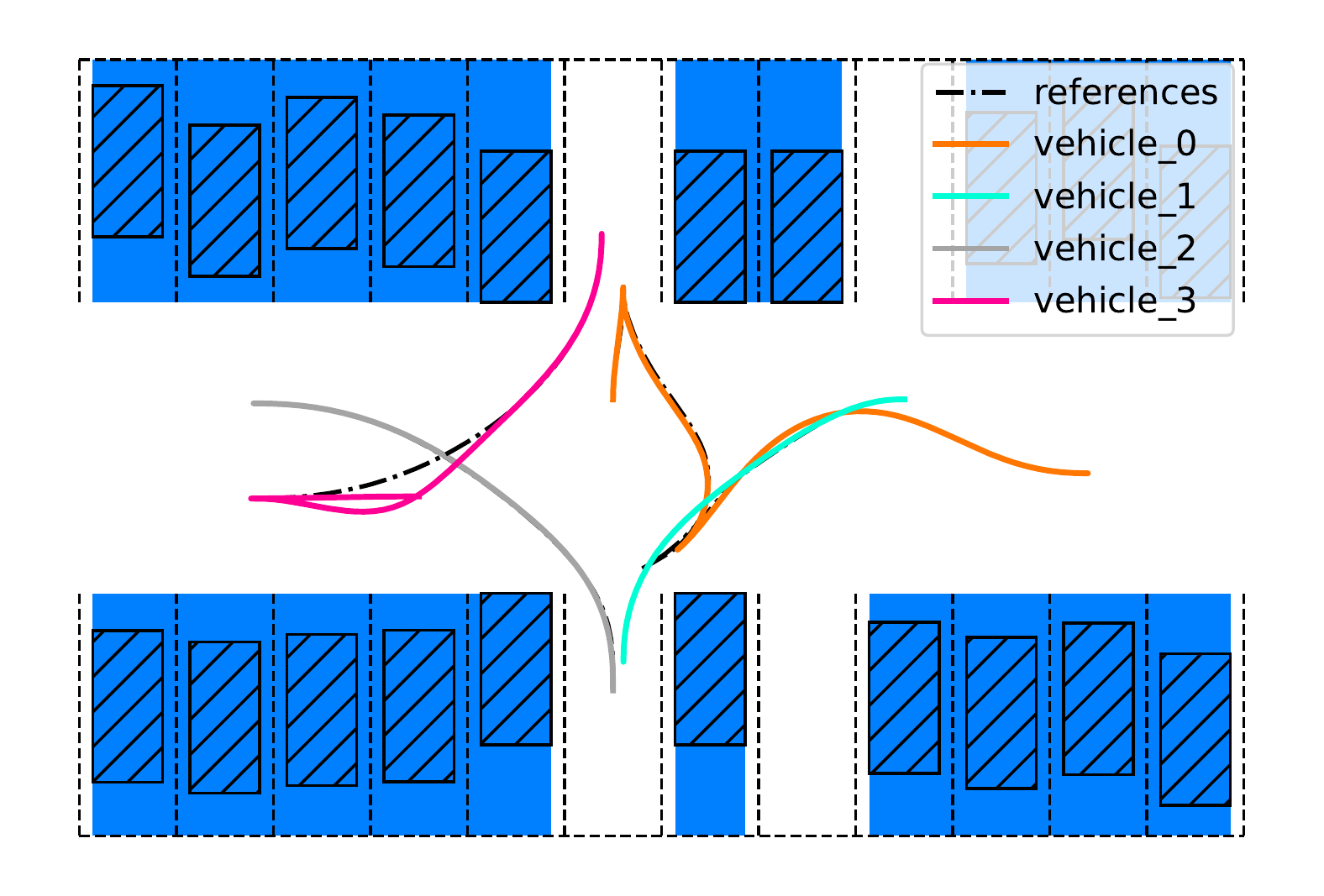}
	    \caption{References and final trajectories}
	    \label{fig:ref-vs-final}
	\end{subfigure}%
	~
	\begin{subfigure}[t]{0.34\columnwidth}
		\centering
		\includegraphics[width=\textwidth]{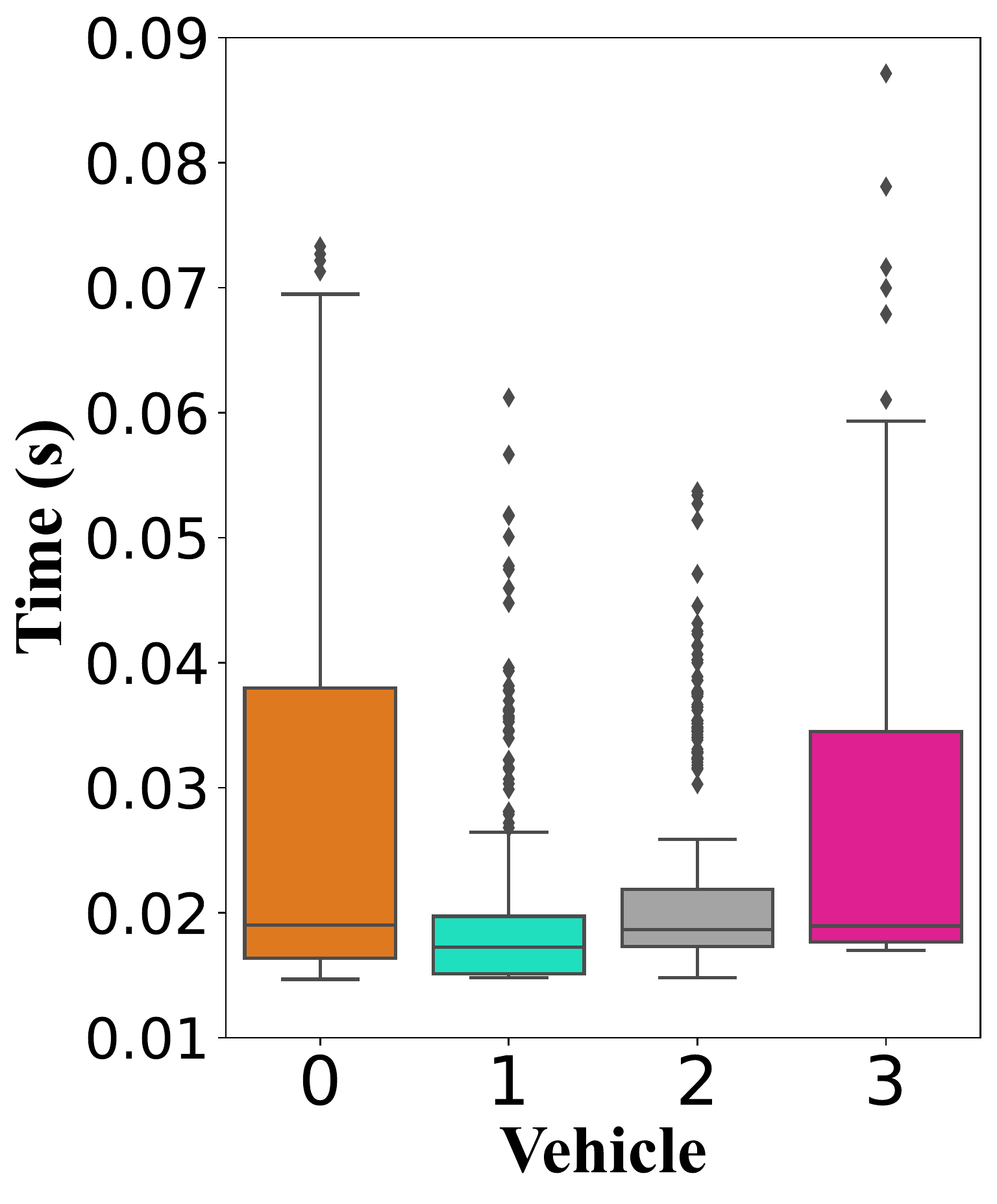}
		\caption{Iteration time}
		\label{fig:time-stats}
	\end{subfigure}
	\caption{Tracking performance and iteration time statistics}
\end{figure}

Once the reference trajectories $\mathbf{z}_{\mathrm{ref}}^{[i]}$ are computed, vehicles solve the distributed MPC problem~
\eqref{eq:distributed-formulation} to track the references as described by line~\ref{algo:for-loop} to line~\ref{algo:apply-control} in Algorithm~\ref{algo:algo}. The sampling time of the discrete-time vehicle dynamics~\eqref{eq:kin_model_dt} is $\tau = 0.1$s, and the MPC look-ahead horizon is $N=30$. The cost function $J^{[i]}$ is defined as
\begin{align*}
    J^{[i]}\left(\mathbf{z}^{[i]}_{\cdot | t}, \mathbf{u}^{[i]}_{\cdot | t} \mid \mathbf{z}_{\mathrm{ref}}^{[i]}\right) = \sum_{k=0}^{N} & \| z^{[i]}_{k|t} - z^{[i]}_{\mathrm{ref}, k|t}\|^{2}_{Q} \\
    \left(\psi^{[i]}_{k|t}\right)^2 & + \left(v^{[i]}_{k|t} w^{[i]}_{k|t}\right)^2 + \left(a^{[i]}_{k|t}\right)^2,
\end{align*}
which reflects the deviation from the reference trajectory, passenger comforts, and the amount of actuation. $\left\{z^{[i]}_{\mathrm{ref}, 0|t}, \dots, z^{[i]}_{\mathrm{ref}, N|t}\right\}$ are sampled along $\mathbf{z}_{\mathrm{ref}}^{[i]}$ and the weight matrix $Q = \mathrm{diag}([100, 100, 100, 0, 0])$.
Fig.~\ref{fig:final-poses} shows some snapshots of vehicle configurations during online control. It can be observed from Fig.~\ref{fig:state-input-profile} that the controller generates smooth speed and steering profiles for vehicles while keeping the inputs within the operating limits. Fig.~\ref{fig:ref-vs-final} compares vehicles' final trajectories with their references, where we can find that vehicles track the references accurately most of the time, except that \textcolor{v3}{vehicle 3} has to deviate temporarily to avoid collision against \textcolor{v2}{vehicle 2}.

We record the computation time for each vehicle to solve problem~\eqref{eq:distributed-formulation} in Fig.~\ref{fig:time-stats}. The data is collected by testing the proposed algorithm on a laptop with quad-core Intel® Core™ i9-9900K CPU @ 3.60GHz. The median iteration time for all vehicles is around 0.02 seconds, and 98.2\% are below 0.1 seconds. The longest iteration among the outliers has a duration of 0.5 seconds.

\section{Conclusion}
\label{sec:conclusion}

This paper proposes a distributed algorithm to resolve conflicts in highly constrained environments, combining deep Multi-Agent Reinforcement Learning (RL) and distributed Model Predictive Control (MPC).

Offline, we can train a policy with deep RL to explore combinatorial actions for vehicles to resolve conflicts in a discrete environment. The trained policy can offer discrete guidance for generating high-quality reference trajectories given a specific scenario. Online, a distributed MPC is formulated to track the reference trajectories while avoiding collision among vehicles. At each time step, the vehicles compute their states and inputs over the look-ahead horizon and communicate their predictions with other vehicles. The simulation results show that the proposed algorithm can control the vehicles in real-time to resolve conflicts safely and efficiently with smooth motion profiles.

% \section*{ACKNOWLEDGMENT}
% We would like to thank Michelle Pan and Dr. Vijay Govindarajan for their contribution in building DLP dataset.
%The preferred spelling of the word  acknowledgment  in America is without an  e  after the  g . Avoid the stilted expression,  One of us (R. B. G.) thanks . . .   Instead, try  R. %B. G. thanks . Put sponsor acknowledgments in the unnumbered footnote on the first page.

% \newpage
\bibliographystyle{IEEEtran}

\bibliography{references.bib}

% \appendix
% \begin{appendices} 
 
% \subsection{Intention prediction module}\label{appendix:goal}
% We use the following parameters in the intention prediction module from Fig.~\ref{fig:lstm_overall} with the architecture:
% \begin{itemize}
%     \item[-] {\tt input0}: $\mathcal{X}_\mathrm{hist}^{(i)}$ with dimension $N_\mathrm{hist}\times{}3$.
%     \item[-] {\tt input1}: $\mathcal{O}^{(i)}$ with dimensions $G\times{}3$.
%     \item[-] {\tt lstm0}: LSTM unit that takes {\tt input0} as input and consists of $100$ hidden units
%     \item[-] {\tt fcc0}: Fully connected unit that takes the output from {\tt lstm0}, concatenated with {\tt input1} as input and consists of $X$ neurons with ReLU activation.
%     \item[-] {\tt output}: Fully connected  unit that takes the output from {\tt fcc0} as input and consists of $33$ neurons with softmax activation.
% \end{itemize}

% \end{appendices}
\end{document}